\documentclass{article}
\usepackage{spconf,amsmath,amssymb,graphicx,caption,subcaption,url}

\title{Directional Bilateral Filters}
\name{Manasij Venkatesh and Chandra Sekhar Seelamantula \thanks{This work is fully supported by the Robert Bosch Center for Cyber Physical Systems, Indian Institute of Science, Bangalore.}}
\address{Department of Electrical Engineering, Indian Institute of Science, Bangalore-560012, India\\
	{\small \tt manasij.ece.nitk@gmail.com, chandra.sekhar@ieee.org}}

\begin{document}
\ninept
\maketitle

\begin{abstract}
We propose a bilateral filter with a locally controlled domain kernel for directional edge-preserving smoothing. Traditional bilateral filters use a range kernel, which is responsible for edge preservation, and a fixed domain kernel that performs smoothing. Our intuition is that orientation and anisotropy of image structures should be incorporated into the domain kernel while smoothing. For this purpose, we employ an oriented Gaussian domain kernel locally controlled by a structure tensor. The oriented domain kernel combined with a range kernel forms the \textit{directional bilateral filter}. The two kernels assist each other in effectively suppressing the influence of the outliers while smoothing. To find the optimal parameters of the directional bilateral filter, we propose the use of Stein's unbiased risk estimate (SURE). We test the capabilities of the kernels separately as well as together, first on synthetic images, and then on real endoscopic images. The directional bilateral filter has better denoising performance than the Gaussian bilateral filter at various noise levels in terms of peak signal-to-noise ratio (PSNR).
\end{abstract}


\section{Introduction}

The goal of smoothing an image is to suppress noise, and emphasize important features. A space-invariant linear filter performs uniform smoothing and is not suitable for preserving edges. Since edges contain the primal sketch of an image, it is desirable to preserve them. Nonlinear filters that are data-adaptive were designed to smooth images without blurring the edges. Anisotropic diffusion described by Perona and Malik \cite{Perona} was first used to achieve edge-preserving smoothing. Subsequently, Aurich and Weule employed nonlinear modifications of Gaussian filters \cite{Aurich}. Tomasi and Manduchi proposed generalized bilateral filters whose range filters suppress outliers to achieve edge preservation \cite{Tomasi}. Elad showed that the bilateral filter and anisotropic diffusion emerge from a Bayesian framework \cite{Elad}.

The bilateral filter $\phi$ is obtained by combining a domain kernel and a range kernel
\begin{equation}
\phi_{\mathbf{p},\mathbf{q}}(y_\mathbf{p},y_\mathbf{q}) = w_{\mathbf{p} - \mathbf{q}}~r(y_\mathbf{p}-y_\mathbf{q}),
\end{equation}
where the domain kernel $w_{\mathbf{p} - \mathbf{q}}$ depends on the geometric distance between the pixel of interest $\mathbf{p}$ and a neighboring pixel $\mathbf{q}$. The kernel is chosen such that the averaging is localized to an $\Omega$-neighbourhood of $\mathbf{p}$. It assigns weights that fall off with decreasing geometric distance. The range kernel $r(y_\mathbf{p}-y_\mathbf{q})$ measures the similarity between the intensity of the pixel of interest $y_\mathbf{p}$ and a neighbourhood pixel $y_\mathbf{q}$. It assigns higher weights to pixels that have relatively similar intensities. The bilateral filter output is
\begin{equation}
\hat{x}_{\mathbf{p}} = h_{\mathbf{p}}^{-1} \sum_{\mathbf{q}\in\Omega}\phi_{\mathbf{p},\mathbf{q}}(y_\mathbf{p},y_\mathbf{q})~y_\mathbf{q},
\end{equation}
where $h_{\mathbf{p}}$ is the normalizing factor given by
\begin{equation}
h_{\mathbf{p}} = \sum_{\mathbf{q}\in\Omega}\phi_{\mathbf{p},\mathbf{q}}(y_\mathbf{p},y_\mathbf{q}).
\end{equation}

\subsection{Related work}

The nonlinearity of $r$ makes the bilateral filter computationally expensive in its standard form. However, they remain attractive as a number of works have been dedicated to accelerate them. Paris and Durand derived criteria for downsampling in space and intensity to come up with a fast approximation of the bilateral filter \cite{Paris}. A constant-time algorithm for fast bilateral filtering has been proposed in \cite{Porikli, Kunal}. Yang et al. achieved substantial acceleration at the cost of quantization \cite{Yang}.

Modifications of the bilateral filters have found widespread use in a number of image processing tasks such as denoising \cite{Bennett}, illumination compensation \cite{Elad2}, optical-flow estimation \cite{Xiao}, demoaiscking \cite{Ramanath}, edge detection \cite{Jose}, etc. 

Considerable work has also been done on optimizing the parameters of the bilateral filter for improving denoising performance. Peng and Rao used Stein's unbiased risk estimate (SURE) to find the optimal parameters of the Gaussian bilateral filter \cite{Peng,Stein}. Kishan and Seelamantula achieved this goal for a bilateral filter with a raised cosine range kernel \cite{Kishan}. Chen and Shu used Chi-square unbiased risk estimate (CURE) for optimizing bilateral filter parameters in squared magnitude MR images \cite{Luisier,Chen}.

\subsection{This paper}

Traditionally, bilateral filters use fixed domain kernels. Our intuition is that if the domain kernel can be locally controlled and adapted to smooth perpendicularly to the dominant orientations in image structures, the influence of outliers can be suppressed while smoothing. 

We propose a domain kernel that can incorporate orientation and anisotropy by means of a structure tensor \cite{Bigun}. This domain kernel is combined with a range kernel to ensure edge preservation. We evaluate SURE for the directional bilateral filter and show that SURE follows the MSE closely. We determine the optimal parameters of this filter by minimizing the SURE cost. We show that considering orientation and anisotropy of image structures, denoising performance is improved.

The paper is organized as follows. We begin with a brief introduction of the Gaussian bilateral filter in Section~\ref{sec:GBF}. The proposed directional bilateral filter, consisting of the anisotropic domain kernel and the range kernel, are detailed in Section~\ref{sec:ABF}. In Section~\ref{sec:sure}, we provide SURE calculations for the directional bilateral filter. The accuracy of SURE and its closeness to MSE are shown in Section~\ref{sec:exp}, along with experimental quantitative and qualitative comparisons with the Gaussian bilateral filter on synthetic images as well as real endoscopic images. Concluding remarks are drawn in Section~\ref{sec:conc}.

\section{The Gaussian Bilateral Filter} \label{sec:GBF}

The Gaussian bilateral filter employs Gaussian domain and range kernels and is given by
\begin{equation}
\phi_{\mathbf{p},\mathbf{q}}^{\mathrm{GBF}}(y_\mathbf{p},y_\mathbf{q}) = \underbrace{\exp\left(-\frac{\|\mathbf{p}-\mathbf{q}\|^2}{2\sigma_{d}^{2}}\right)}_{\text{domain kernel}} \underbrace{\exp\left(-\frac{\vert y_\mathbf{p}-y_\mathbf{q} \vert ^{2}}{2\sigma_{r}^{2}}\right)}_{\text{range kernel}}.
\end{equation}
The domain kernel does not incorporate local orientation and anisotropy measures of image structures and is determined a priori for a given $\sigma_d$. The parameters $\sigma_{d}$ and $\sigma_{r}$ control the rates at which the Gaussian functions decay. Selecting them optimally is crucial for efficient denoising.

\section{Directional Bilateral Filter} \label{sec:ABF} 

\subsection{Anisotropic domain filter} \label{sec:ADF}

The standard Gaussian domain kernel is symmetric around the center of the window $\mathbf{p}$. Since we want to incorporate orientation and anisotropy of image structures while smoothing, we use an oriented Gaussian domain kernel. The anisotropic domain filter (ADF) given by
\begin{equation}
\phi_{\mathbf{p},\mathbf{q}}^{\mathrm{ADF}}(y_\mathbf{p},y_\mathbf{q}) = \exp\left(-\frac{\gamma_1^2 m^2 + \gamma_2^2 n^2}{2\rho_{d}^{2}} \right),
\end{equation}
where 
\begin{equation*}
\begin{array}{lclcl}
m &=& ~~~(m_\mathbf{q} - m_\mathbf{p})\cos\theta &+& (n_\mathbf{q} - n_\mathbf{p})\sin\theta, \mathrm{~and}\\
n &=& -(m_\mathbf{q} - m_\mathbf{p})\sin\theta &+& (n_\mathbf{q} - n_\mathbf{p})\cos\theta;
\end{array}
\end{equation*}
($m_\mathbf{p},n_\mathbf{p}$), ($m_\mathbf{q},n_\mathbf{q}$) are the coordinates of pixels $\mathbf{p}$ and $\mathbf{q}$.

The additional parameters $\gamma_1$, $\gamma_2$, and $\theta$ control the scaling and the orientation of the oriented Gaussian. They allow smoothing along a particular direction by taking into consideration the orientation and anisotropy. We locally obtain the $\gamma$ and $\theta$ parameters using the structure tensor approach.

\subsection{Structure tensor}

A structure tensors gives accurate orientation estimation and local anisotropy measures in neighbourhoods. Let the grayscale image be denoted by $I$. The difference of Gaussians (DoG) kernel is used to compute the gradient of the image $\nabla I$. The 2-D structure tensor $J_{\rho}$, is a smoothed version of the second moment matrix $(\nabla I) (\nabla I)^T $. The smoothing is performed by convolving the matrix components with a Gaussian kernel $G_{\rho}$ with standard deviation ${\rho}$:
\begin{equation}
\begin{array}{lclcl}
	J_{\rho} & = &  \begin{bmatrix}  G_{\rho} \ast I_{x}I_{x}~~~~G_{\rho} \ast I_{x}I_{y}\\G_{\rho} \ast I_{x}I_{y}~~~~G_{\rho} \ast I_{y}I_{y} \end{bmatrix} & = & \begin{bmatrix} J_{11}~~J_{12}\\J_{21}~~J_{22} \end{bmatrix}.\\
\end{array}	
\end{equation}

By construction, $J_{\rho_d}$ is a symmetric, positive semidefinite matrix. The information about orientation and anisotropy is obtained by eigen-value decomposition. 

The eigenvalues are obtained directly from $J_{\rho}$ as:
\begin{equation}
\lambda_{1,2} = \frac{1}{2}\left(J_{22} + J_{11} \pm \sqrt{(J_{22} - J_{11})^2 + 4J_{12}^2}\right).
\end{equation}
The parameters $\theta$ and $\gamma$ are obtained as follows:
\begin{enumerate}
\item We orient the domain kernel perpendicular to the direction of the dominant orientation:
\begin{equation}
\theta = \frac{\pi}{2} + \tan^{-1}\left(\frac{2J_{12}}{J_{22}-J_{11}}\right).
\end{equation}
\item Big\"un and Granlund \cite{Bigun} described a certainty measure $C$ as
\begin{equation*}
C = \left(\frac{\lambda_2 - \lambda_1}{\lambda_2 + \lambda_1}\right),
\end{equation*}
which is a measure of anisotropy. We set the scaling factors as
\begin{equation}
\gamma_2 = (1+C) \mathrm{~~~and~~~} \gamma_1 = 1/\gamma_2.
\end{equation}
In constant neighbourhoods, where $\lambda_1 + \lambda_2 = 0$, $C$ is set to 0 since there is no unique orientation. The aspect ratio increases with increasing anisotropy.
\end{enumerate}

The structure tensor contains no more information than the gradient itself but has the distinct advantage that the matrix can be smoothed without cancellation effects in areas where gradients have opposite signs, since $(\nabla I) (\nabla I)^T = (-\nabla I)(-\nabla I)^T$.

The oriented Gaussian kernel adapts to the data according to $\gamma$ and $\theta$ obtained from the structure tensor and helps to smooth along edges. Furthermore, a range kernel is used along with the oriented Gaussian domain kernel to assist in edge preservation. The directional bilateral filter (DBF) is given by
\begin{equation} \label{eq:dbf}
\phi_{\mathbf{p},\mathbf{q}}^{\mathrm{DBF}}(y_\mathbf{p},y_\mathbf{q}) = \exp\left(-\frac{\gamma_1^2 m^2 + \gamma_2^2 n^2}{2\rho_{d}^{2}} \right)\exp\left(-\frac{\vert y_\mathbf{p}-y_\mathbf{q} \vert ^{2}}{2\rho_{r}^{2}}\right).
\end{equation}

\section{Optimal Parameters of the Directional Bilateral Filter using SURE} \label{sec:sure}

Consider an image $\mathbf{x}$ (vector representation of an image) corrupted by additive white Gaussian noise $\mathbf{n}$ of zero-mean and $\sigma^2$I covariance matrix. The noise image $\mathbf{y}$ is given by $\mathbf{y} = \mathbf{x} + \mathbf{n}$. The denoised image $\hat{\mathbf{x}}$ should be an accurate estimate of $\mathbf{x}$. The MSE, which quantifies the closeness of the filtered image to the original, is defined as $\mathrm{MSE}(\hat{\mathbf{x}}) = \mathcal{E}\{\|\hat{\mathbf{x}} - \mathbf{x}\|^2\}$. For a given image, it is expressed as 
\begin{equation} \label{eq:risk}
\mathrm{MSE}(\hat{\mathbf{x}}) = \frac{1}{N}\|\hat{\mathbf{x}} - \mathbf{x}\|^2,
\end{equation}
where $N$ is the total number of pixels in the image. In a practical scenario, we do not have access to the original image $\mathbf{x}$. We propose to use SURE to obtain the optimal parameters $\rho_d, \rho_r$ of the DBF.

\begin{figure*}[t]
	\begin{center}
		$\begin{array}{ccccc}
		\includegraphics[width=1.1in]{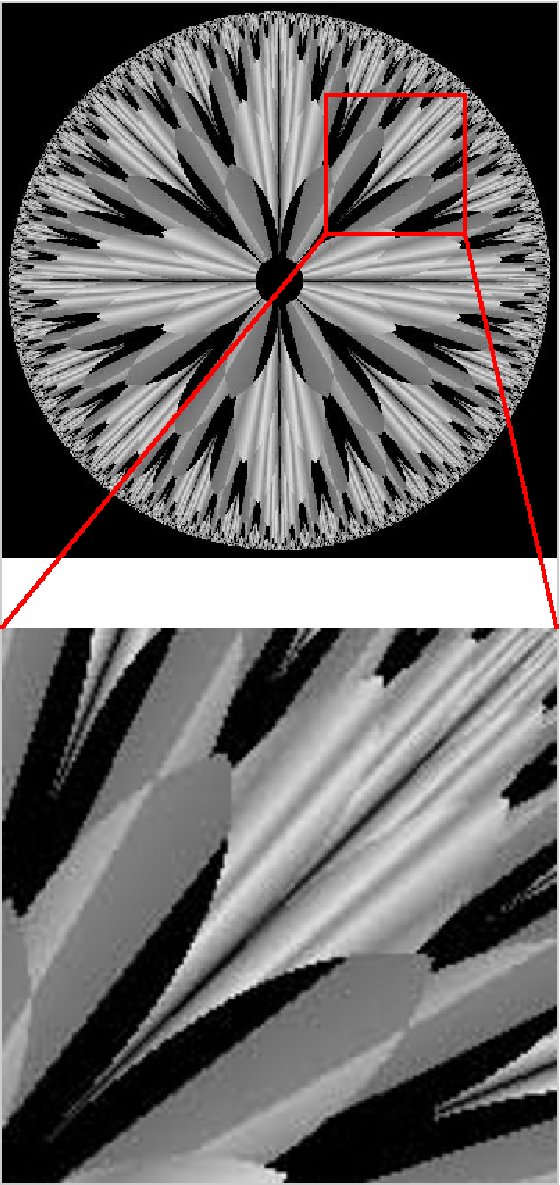}~~~ &
		\includegraphics[width=1.1in]{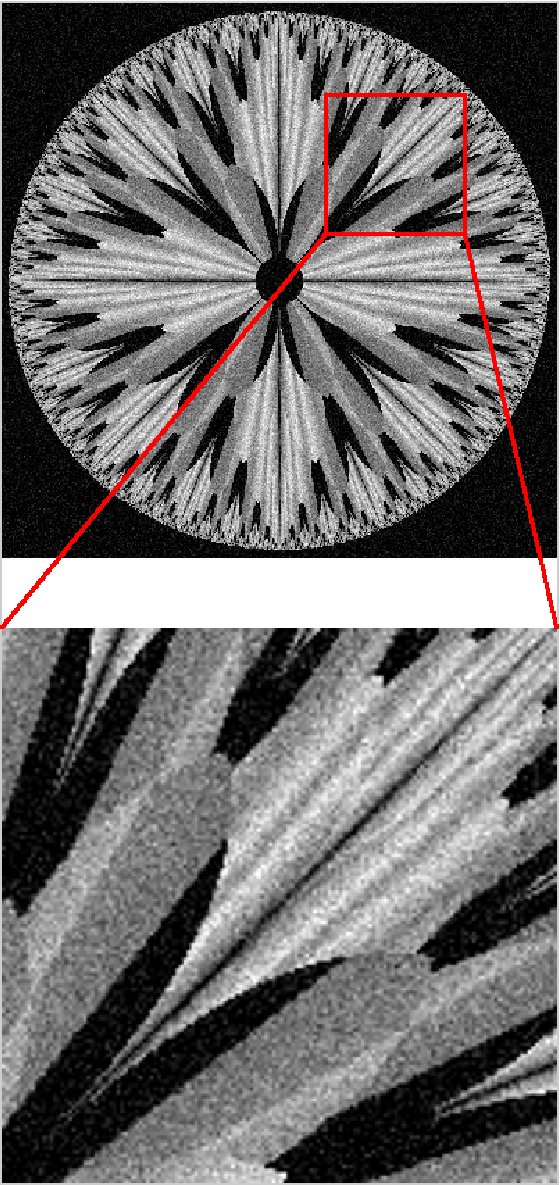}~~~ &
		\includegraphics[width=1.1in]{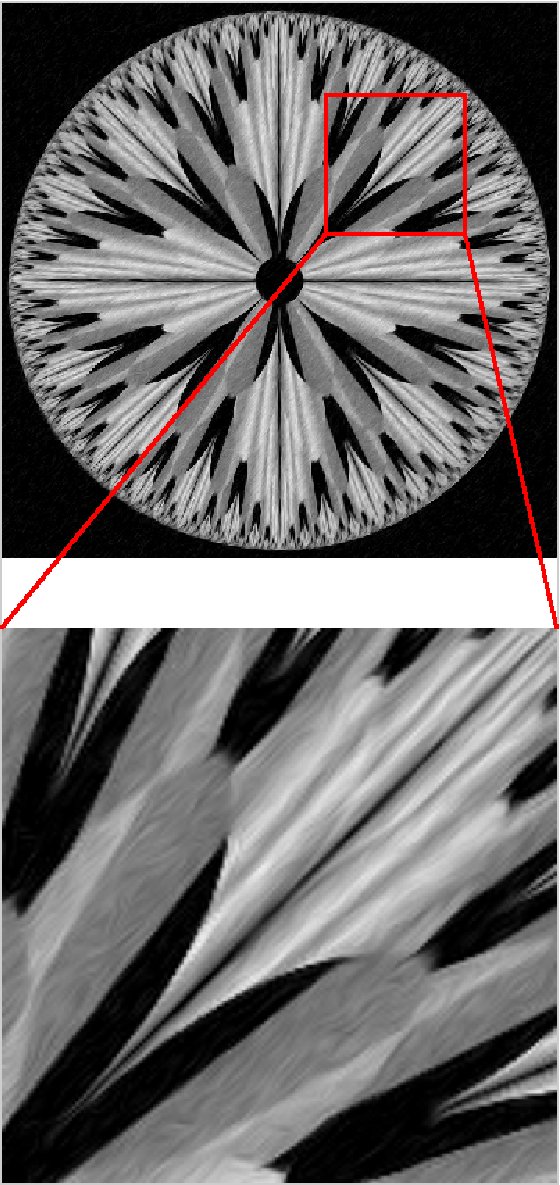}~~~ &
		\includegraphics[width=1.1in]{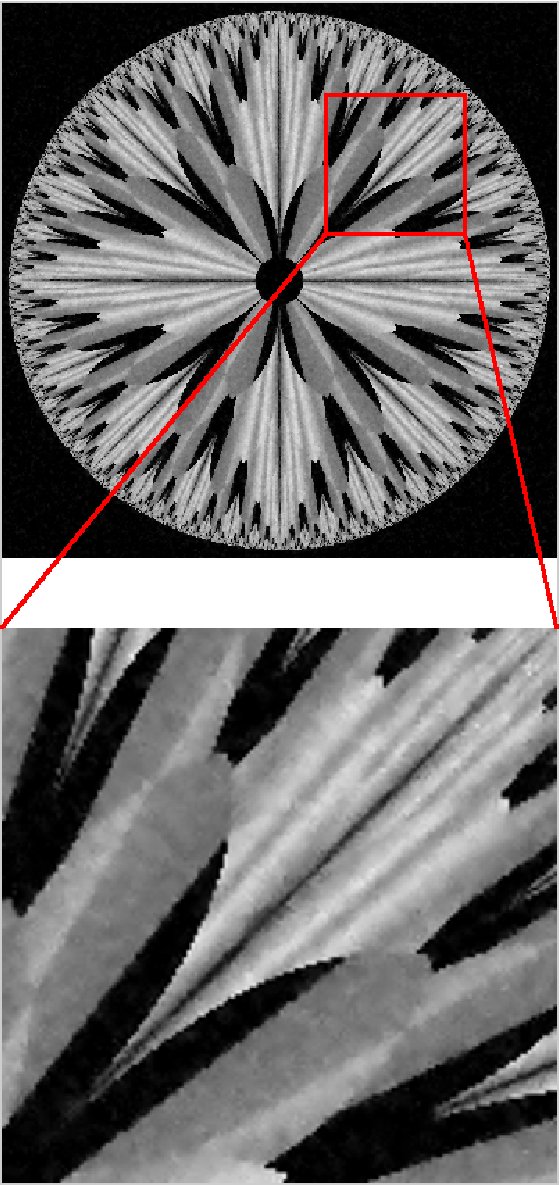}~~~ &
		\includegraphics[width=1.1in]{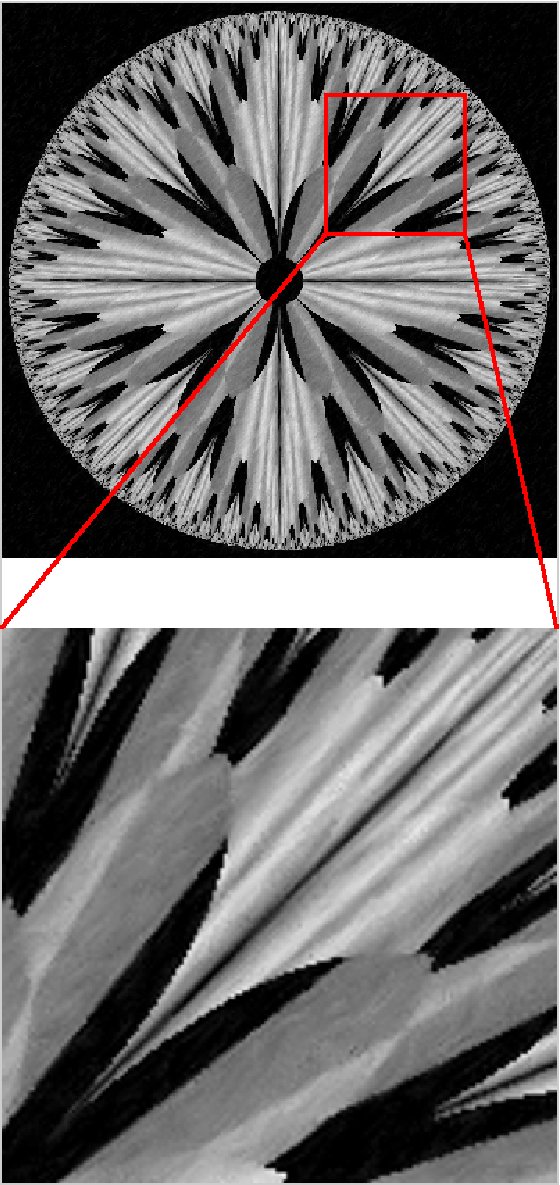}~~~\\
		\mbox{(a)}&\mbox{(b)}&\mbox{(c)}&\mbox{(d)}&\mbox{(e)}\\
		\end{array}$
	\end{center}
	\vspace{-0.25in}
	\caption{Filtered images and their zoomed versions with PSNR values indicated: (a) Original image \cite{Gonzalez}, (b) Noisy (21.80 dB), (c) Proposed anisotropic domain filter (26.60 dB), (d) Gaussian bilateral filter (26.95 dB), and (e) Proposed directional bilateral filter (27.75 dB).}
	\label{fig:fringe3}
\end{figure*}

From \cite{Peng,Stein}, an unbiased estimate of (\ref{eq:risk}) is given by
\begin{equation}
\mathrm{SURE}(\hat{\mathbf{x}}) = \frac{1}{N}\|\hat{\mathbf{x}} - \mathbf{y}\|^2 + \frac{2\sigma^2}{N} \mathrm{div_y}(\hat{\mathbf{x}}) - \sigma^2.
\end{equation}
The divergence term is given by
\begin{equation} \label{eq:div}
\mathrm{div_y}(\hat{\mathbf{x}}) = \sum_{p \in I}\frac{\partial \hat{x}_p}{\partial y_p}.
\end{equation}
\begin{figure}[t]
	\begin{center}
		$\begin{array}{cc}
		\includegraphics[width=1.6in]{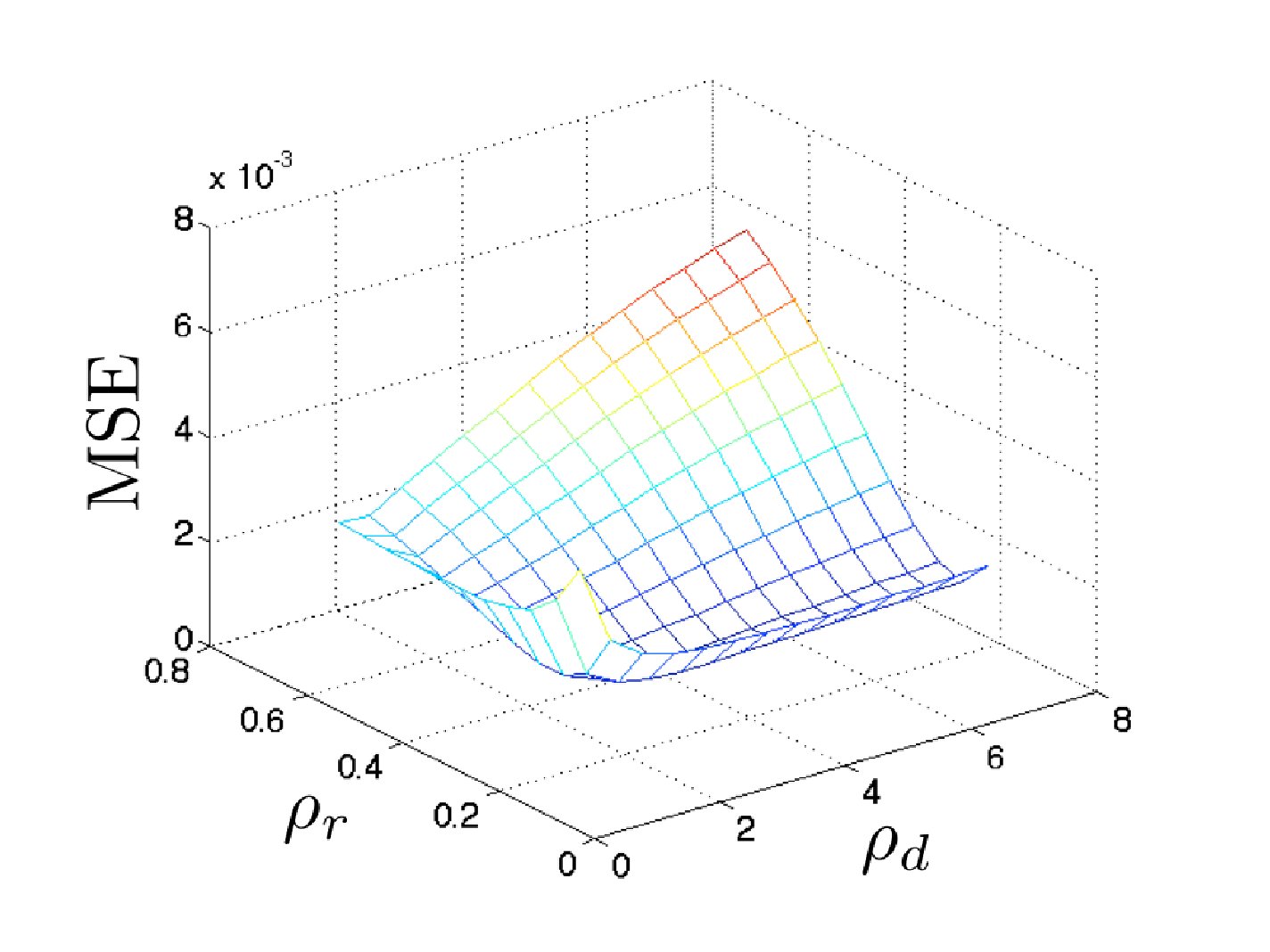} &
		\includegraphics[width=1.6in]{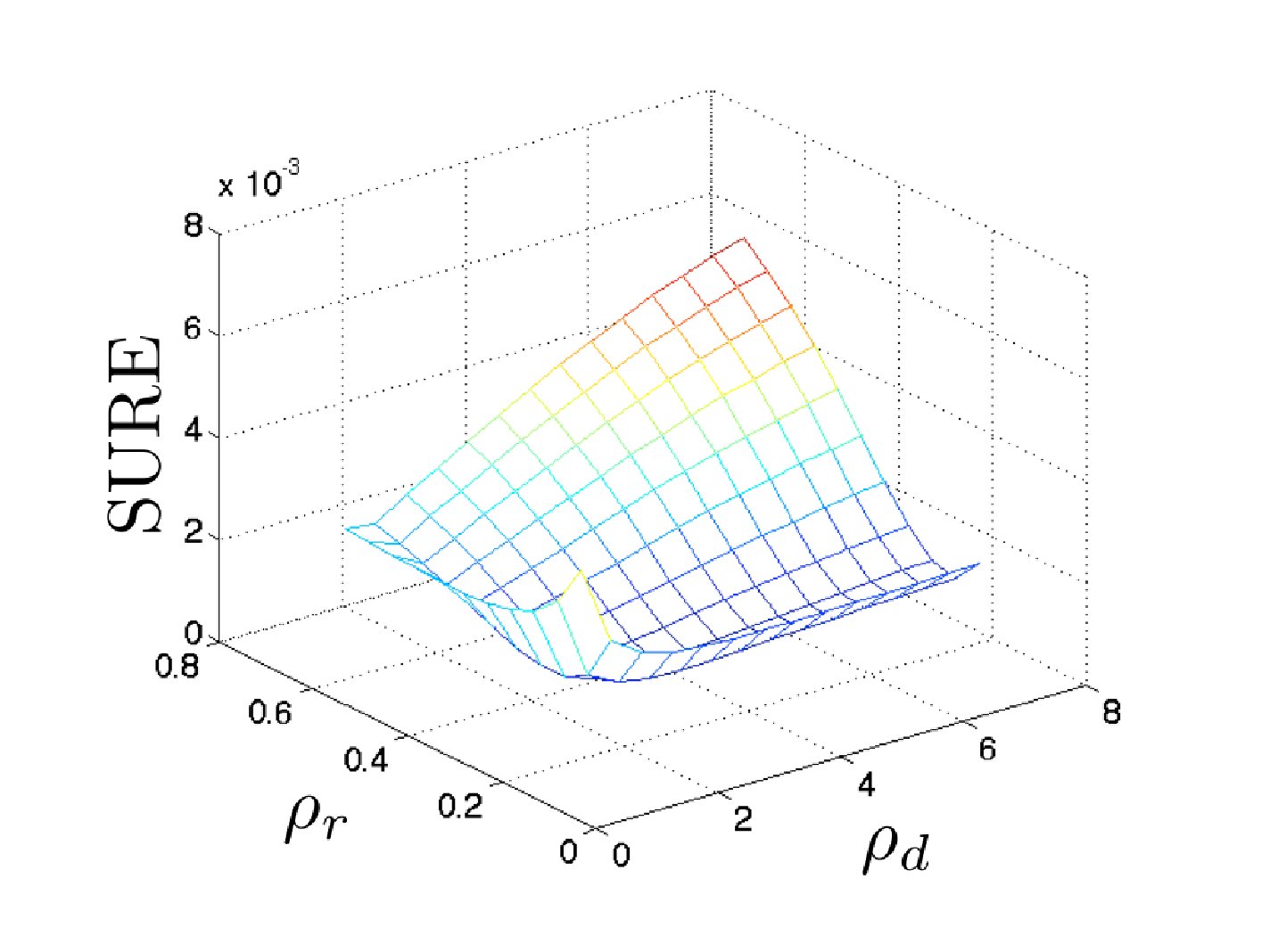} \\
		\mbox{(a)}&\mbox{(b)}\\
		\includegraphics[width=1.6in]{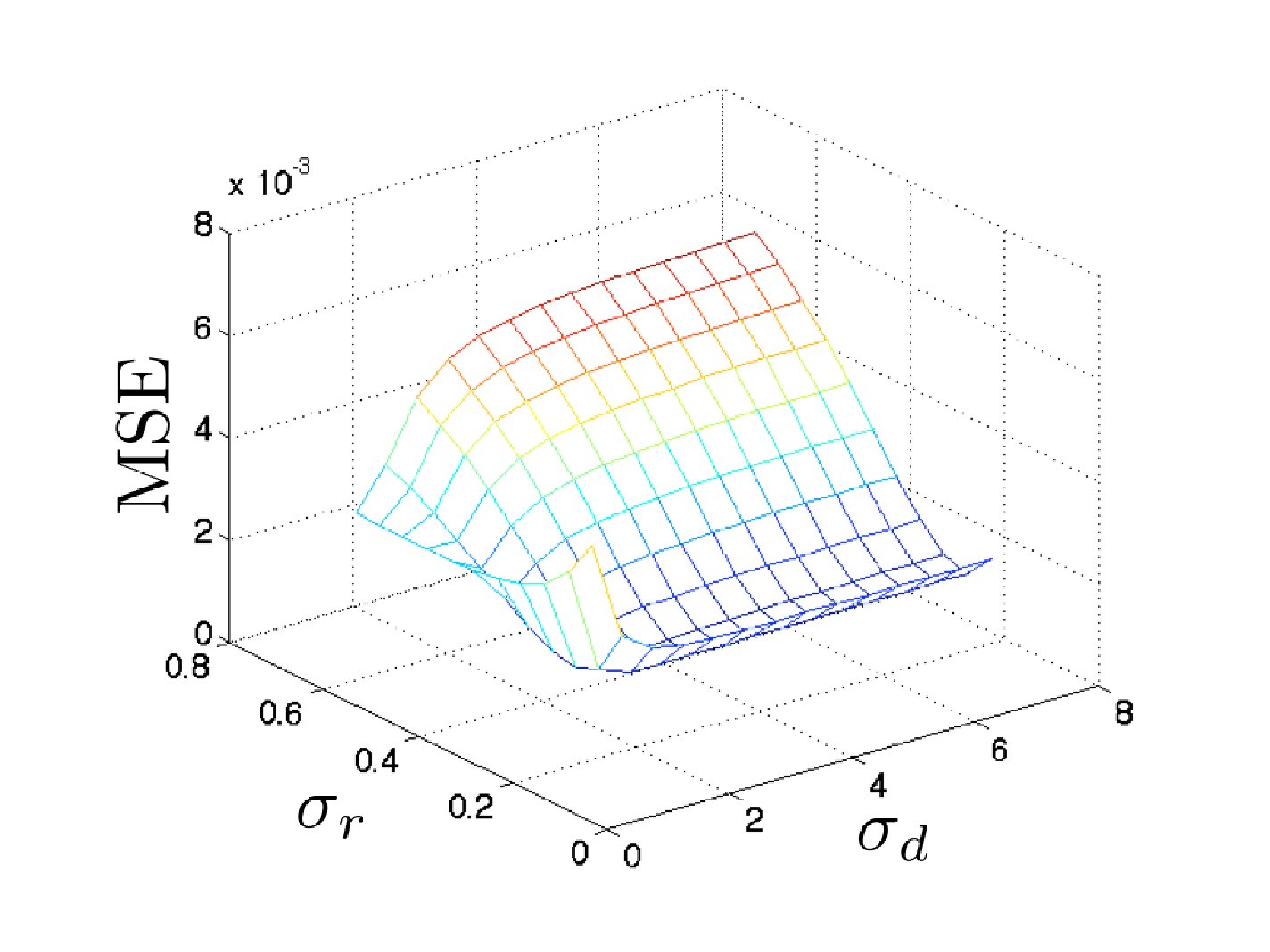} &
		\includegraphics[width=1.6in]{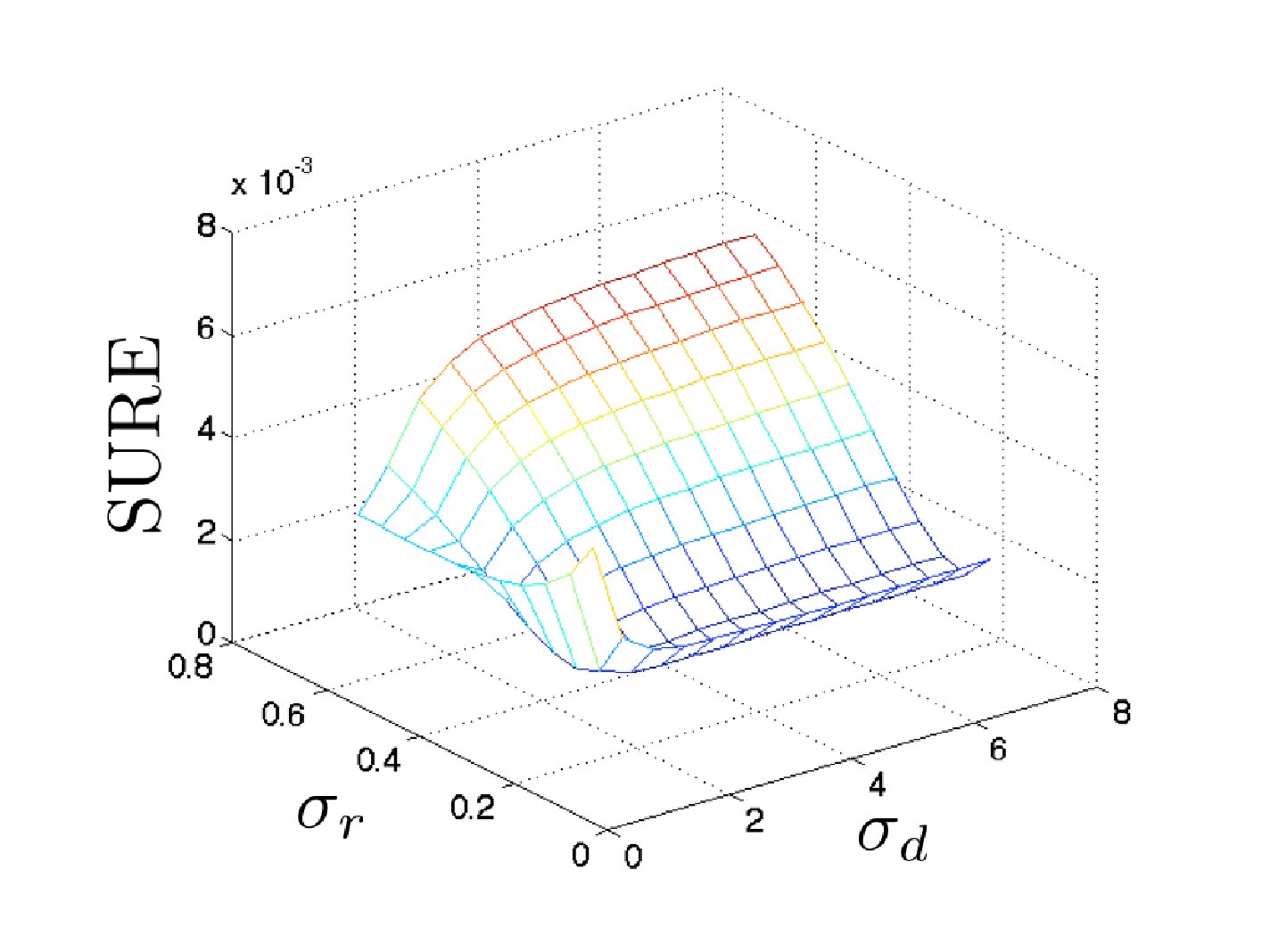}\\
		\mbox{(c)}&\mbox{(d)}\\
		\end{array}$
	\end{center}
	\caption{(color online) Comparison of MSE and SURE plots for the synthetic image. (a) and (b) correspond to the directional bilateral filter, (c) and (d) correspond to the Gaussian bilateral filter. We observe that SURE in (b) and (d) closely approximates the MSE in (a) and (c), respectively.}
	\label{fig:sure}
\end{figure}

\noindent The differential of the filter output with respect to the noisy image is obtained as
\begin{equation*} \label{eq:part1}
\!\!\!\!\!\!\!\!\!\!\frac{\partial \hat{x}_{\mathbf{p}}}{\partial y_{\mathbf{p}}} = h_{\mathbf{p}}^{-1}\left(1 + \sum_{\mathbf{q}\in\Omega}\frac{\partial\phi_{\mathbf{p},\mathbf{q}}^{\mathrm{DBF}}(y_\mathbf{p},y_\mathbf{q})}{\partial y_{\mathbf{p}}}~y_{\mathbf{p}} \right.
\end{equation*}
\begin{equation}
~~~~~~~~~~~~~~~~~~~~~~~~~~~~~~\left. - \hat{x}_p~\sum_{\mathbf{q}\in\Omega}\frac{\partial\phi_{\mathbf{p},\mathbf{q}}^{\mathrm{DBF}}(y_\mathbf{p},y_\mathbf{q})}{\partial y_p}\right).
\end{equation}
Since the domain kernel weights are precomputed, the derivative of (\ref{eq:dbf}) with respect to $y_p$ is
\begin{equation} \label{eq:part2}
\frac{\partial\phi_{\mathbf{p},\mathbf{q}}^{\mathrm{DBF}}(y_\mathbf{p},y_\mathbf{q})}{\partial y_{\mathbf{p}}} = \phi_{\mathbf{p},\mathbf{q}}^{\mathrm{DBF}}(y_\mathbf{p},y_\mathbf{q}) \left( \frac{y_{\mathbf{q}} - y_{\mathbf{p}}}{\sigma_r^2} \right).
\end{equation}
Using (\ref{eq:div}), (\ref{eq:part1}), (\ref{eq:part2}) we evaluate the divergence term and calculate $\mathrm{SURE}(\hat{\mathbf{x}})$.

\begin{table}[!b]
	\caption{Performance comparison of bilateral filter variants in terms of PSNR. Output PSNRs have been averaged over 20 noise realizations. The PSNR values (in dB) are shown.} \label{tab:testimages}
	\vspace{-0.2in}
	\begin{center}
		\begin{tabular}{| c || c  c  c  c  c |}
			\hline
			& \multicolumn{5}{|c|}{{Synthetic image} (600 $\times$ 600)} \\ 
			Input PSNR   & 27.82 & 21.80 & 18.28 & 15.78 & 13.83 \\ \hline
			GBF   	     & 31.59 & 26.95 & 24.05 & 22.32 & 21.14 \\ 
			Proposed ADF & 30.72 & 26.60 & 23.37 & 22.12 & 20.84 \\ 
			Proposed DBF & \textbf{32.38} & \textbf{27.74} & \textbf{24.62} & \textbf{22.61} & \textbf{21.29} \\ \hline
			& \multicolumn{5}{|c|}{{Endoscopy image} in Fig.~\ref{fig:endo} (370 $\times$ 370)} \\ 
			Input PSNR   & 28.12 & 22.07 & 18.56 & 16.03 & 14.16 \\ \hline
			GBF 		 & 37.50 & 33.86 & 31.03 & 28.97 & 26.23 \\ 
			Proposed ADF & 37.24 & 33.89 & 31.26 & 29.49 & 26.76 \\ 
			Proposed DBF & \textbf{38.45} & \textbf{34.20} & \textbf{31.94} & \textbf{30.03} & \textbf{27.26} \\ \hline
			& \multicolumn{5}{|c|}{{Endoscopy image} in Fig.~\ref{fig:endo2} (370 $\times$ 370)} \\ 
			Input PSNR   & 28.13 & 22.11 & 18.60 & 16.10 & 14.15 \\ \hline
			GBF 		 & 38.69 & 34.85 & 31.76 & 29.24 & 26.81 \\ 
			Proposed ADF & 37.41 & 34.63 & 31.35 & 29.51 & 27.12 \\ 
			Proposed DBF & \textbf{39.14} & \textbf{35.39} & \textbf{32.90} & \textbf{30.57} & \textbf{28.00} \\
			\hline
		\end{tabular}
	\end{center}
\end{table}

We compute the optimal parameters of the DBF by minimizing SURE over several values of the parameters $\rho_d$ and $\rho_r$. The optimal parameters of the ADF and GBF are found similarly.

\section{Experimental Results} \label{sec:exp}

\begin{figure*}[t!]
	\begin{center}
		$\begin{array}{ccccc}
		\includegraphics[width=1.05in]{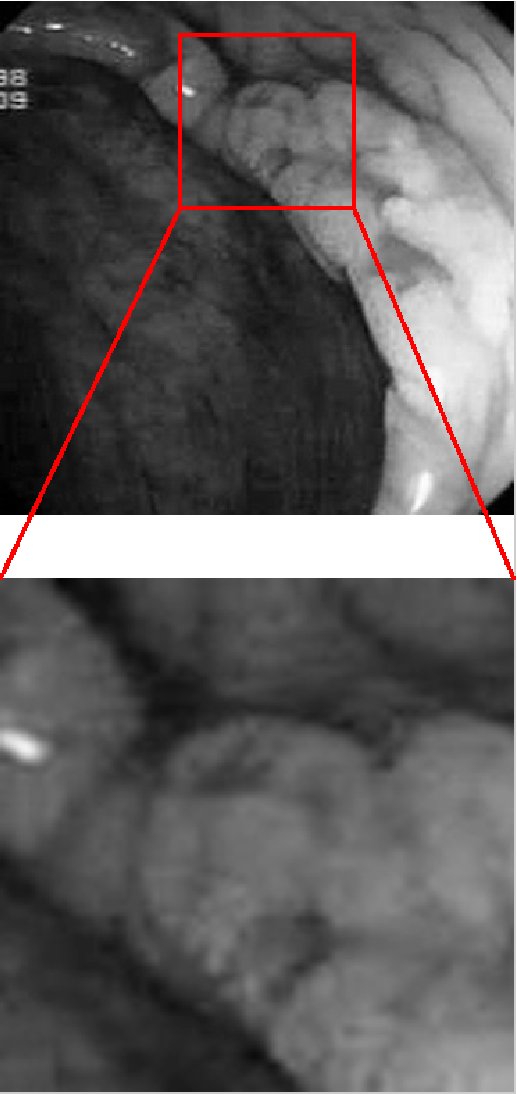}~~~ &
		\includegraphics[width=1.05in]{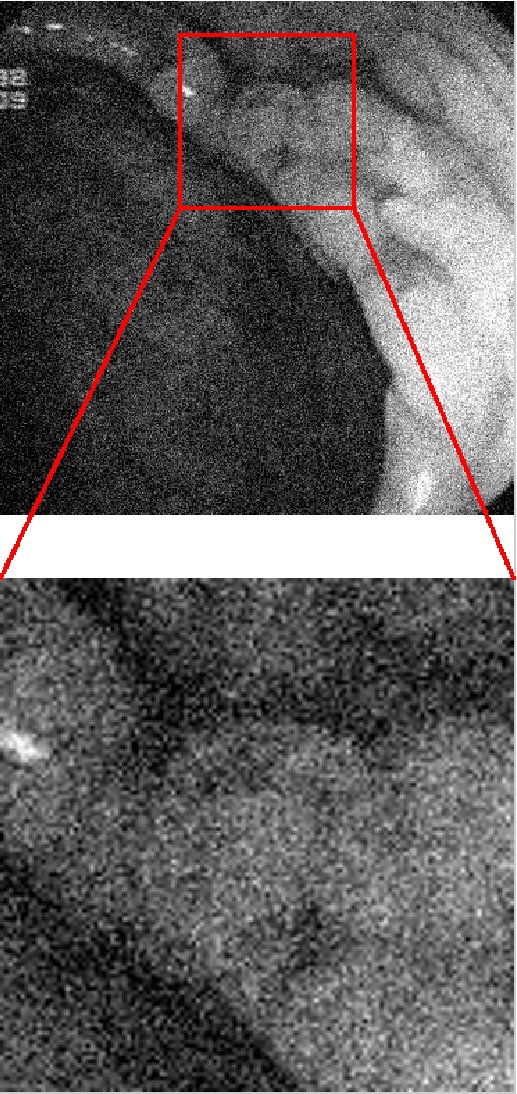}~~~ &
		\includegraphics[width=1.05in]{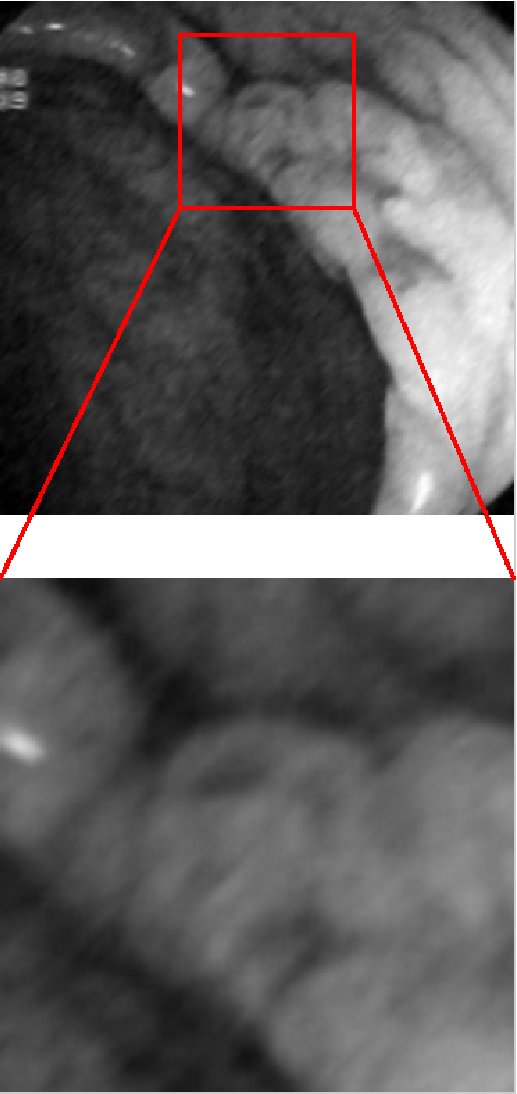}~~~ &
		\includegraphics[width=1.05in]{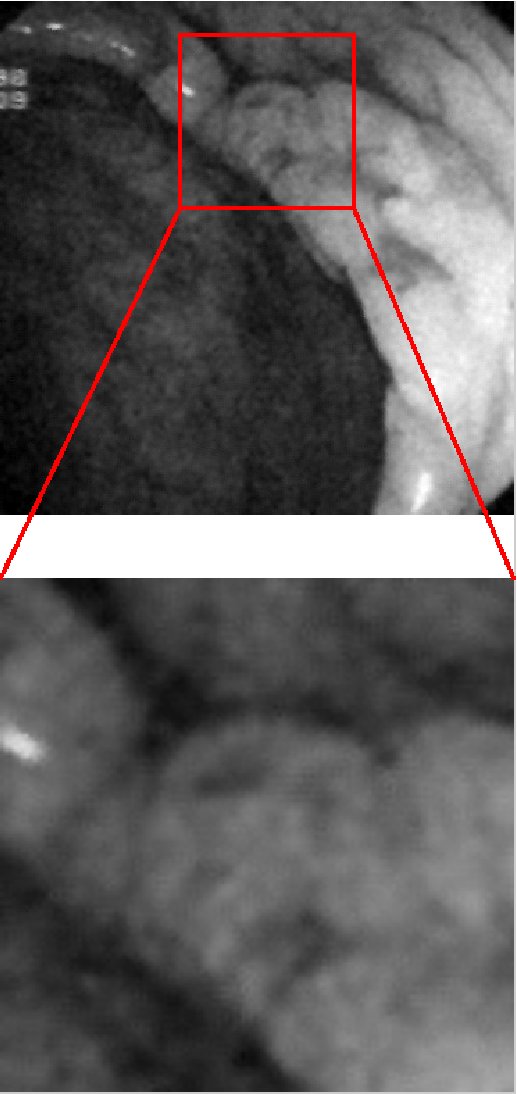}~~~ &
		\includegraphics[width=1.05in]{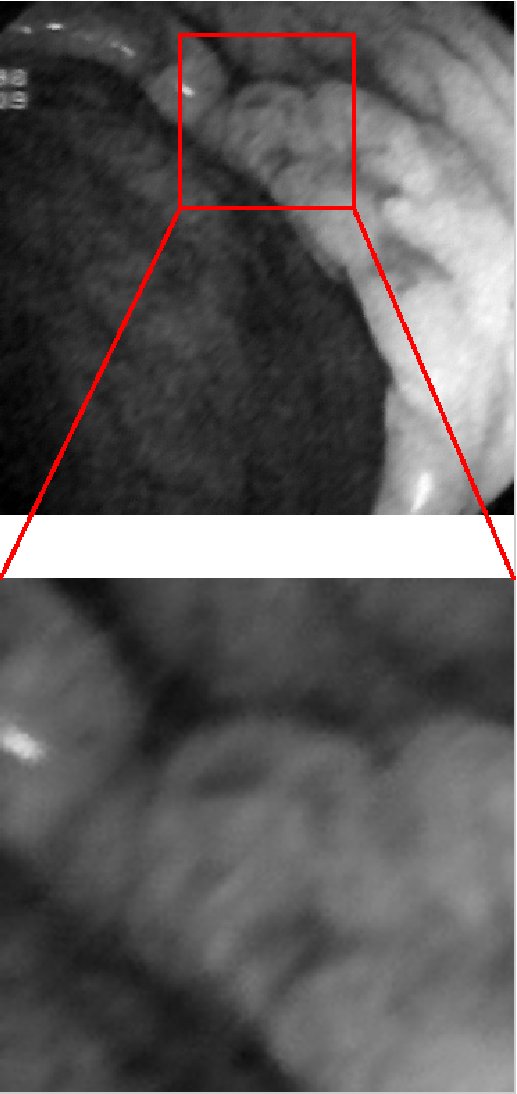}\\
		\mbox{(a)}&\mbox{(b)}&\mbox{(c)}&\mbox{(d)}&\mbox{(e)}\\
		\end{array}$
	\end{center}
	\vspace{-0.25in}
	\caption{Lesions with irregular margins observed in colonoscopy. (Example 1) (a) Original image \cite{web}, (b) Noisy (22.07 dB), (c) Proposed anisotropic domain filter (33.89 dB), (d) Gaussian bilateral filter (33.86 dB), and (e) Proposed directional bilateral filter (34.20 dB).}
	\label{fig:endo}
\end{figure*}

\begin{figure*}[t!]
	\begin{center}
		$\begin{array}{ccccc}
		\includegraphics[width=1.05in]{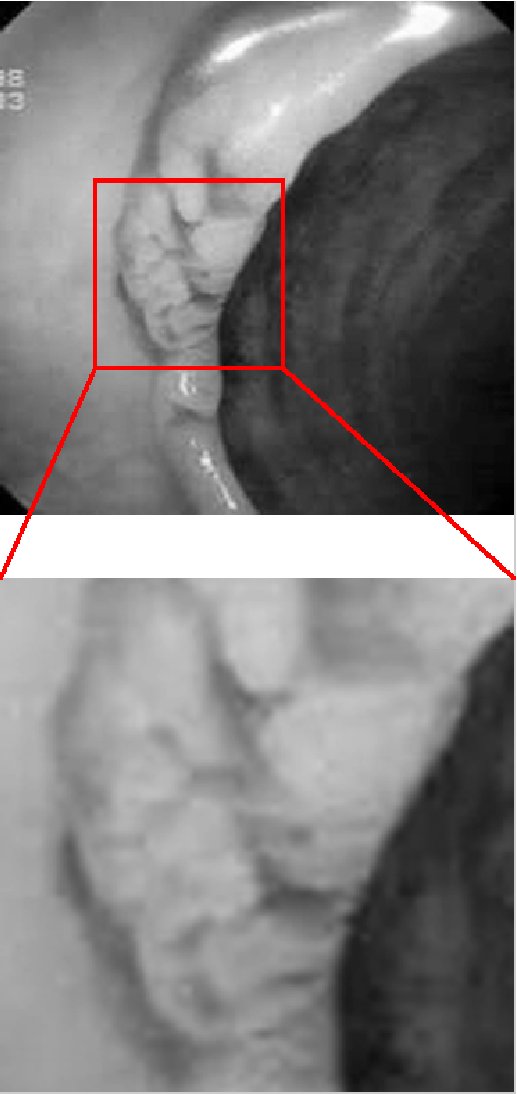}~~~ &
		\includegraphics[width=1.05in]{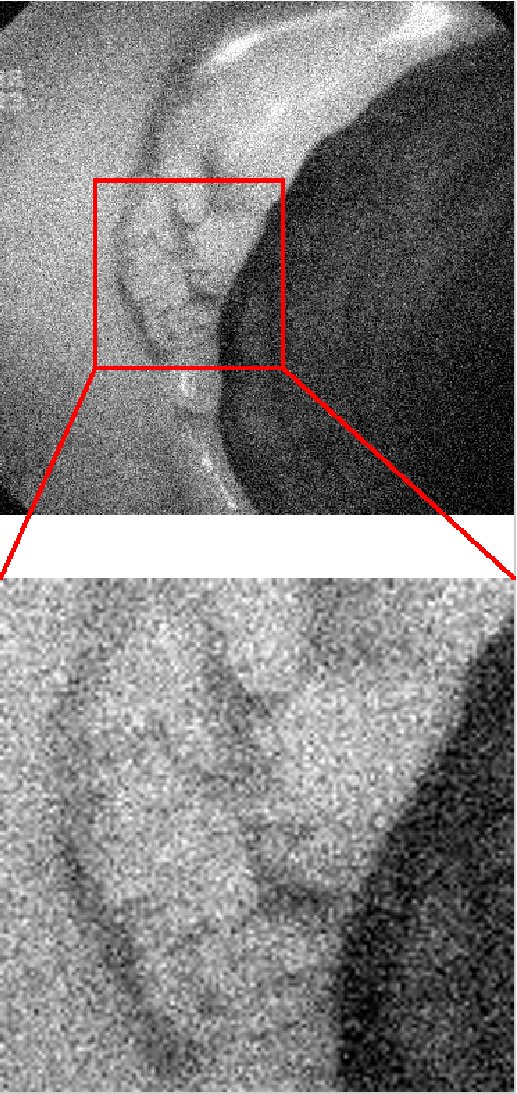}~~~ &
		\includegraphics[width=1.05in]{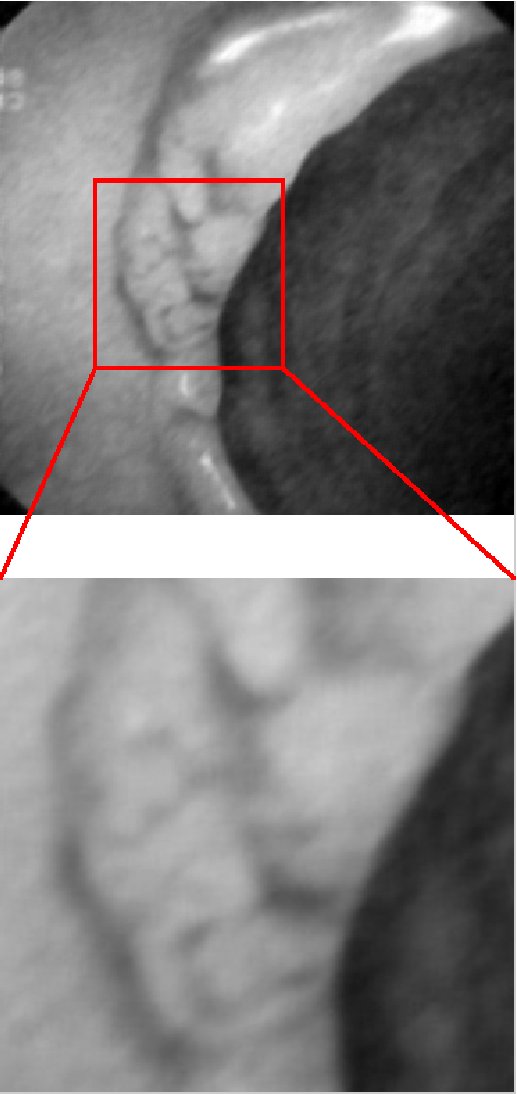}~~~ &
		\includegraphics[width=1.05in]{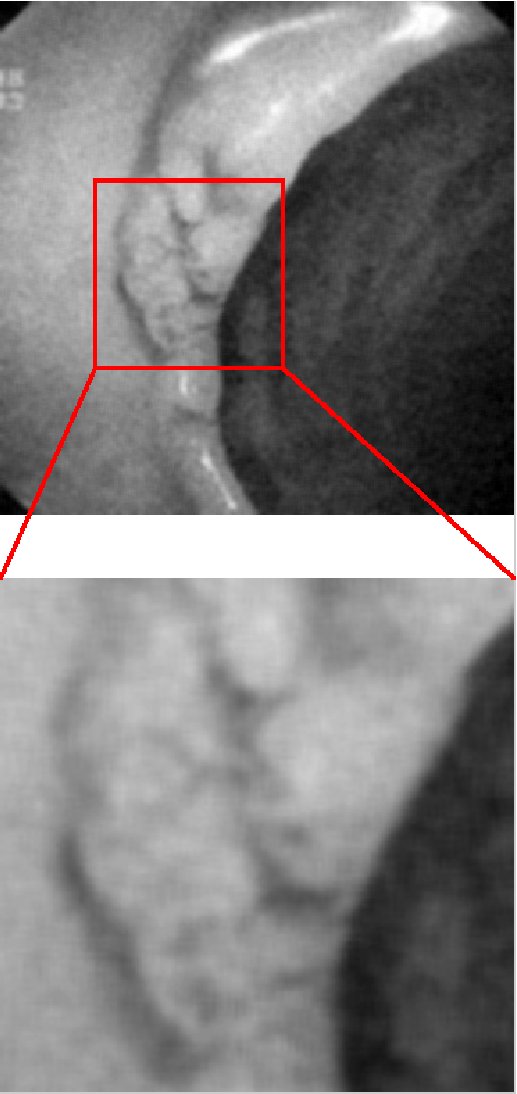}~~~ &
		\includegraphics[width=1.05in]{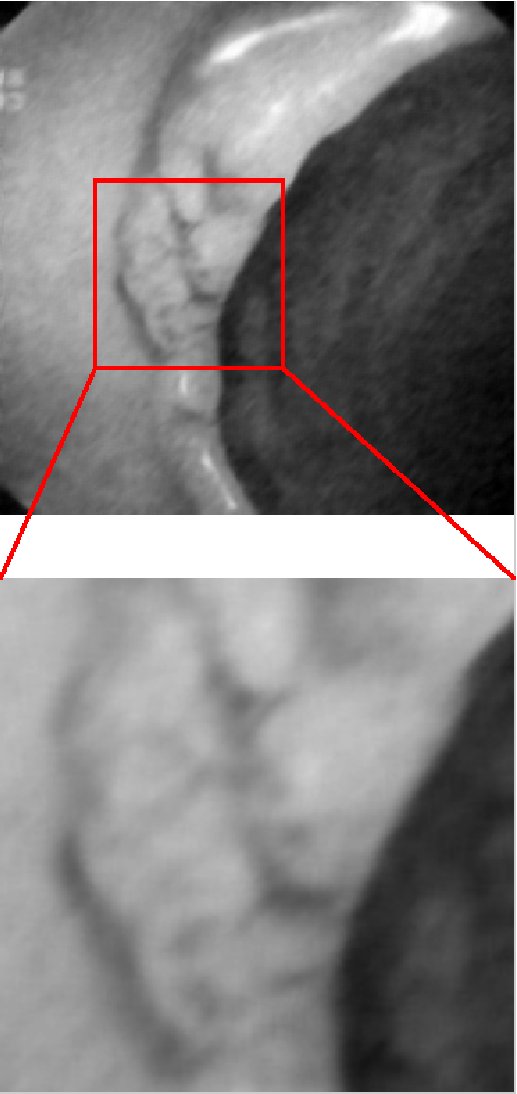}\\
		\mbox{(a)}&\mbox{(b)}&\mbox{(c)}&\mbox{(d)}&\mbox{(e)}\\
		\end{array}$
	\end{center}
	\vspace{-0.25in}
	\caption{Lesions with irregular margins observed in colonoscopy. (Example 2) (a) Original image \cite{web}, (b) Noisy (22.11 dB), (c) Proposed anisotropic domain filter (34.63 dB), (d) Gaussian bilateral filter (34.85 dB), and (e) Proposed directional bilateral filter (35.39 dB).}
	\label{fig:endo2}
\end{figure*}

We first examine if SURE follows the MSE for the directional bilateral filter. We present qualitative and quantitative comparisons between the directional bilateral filter and the Gaussian bilateral filter. For the synthetic image in Fig.~\ref{fig:fringe3}, a noisy realization is obtained by adding zero-mean white Gaussian noise (PSNR = 21.78 dB). The orientation and scaling parameters are computed from the noisy image using the structure tensor approach. The image is then denoised using the directional bilateral filter for different parameter settings $\rho_d$ and $\rho_r$. In each case, the MSE and SURE were computed. The results are shown in Fig.~\ref{fig:sure}. We observe that the SURE closely approximates the MSE. Further, the MSE curve is relatively fat around the optimal parameters for the DBF compared to the GBF. This makes it possible to use a coarser grid of parameters for the search.

In the zoomed versions of the synthetic image (Fig.~\ref{fig:fringe3}), we observe that the edges are preserved better by the DBF (Fig.~\ref{fig:fringe3}(e)) than the GBF (Fig.~\ref{fig:fringe3}(d)). The edges appear sharper because of the directional smoothing. Quantitative comparisons are made based on PSNR measures at various noise levels (Table~\ref{tab:testimages}). The optimal parameters for all the filters are chosen by minimizing the SURE cost. The directional bilateral filter outperforms the Gaussian bilateral filter at all noise levels. We observe that ADF, which does not contain a range kernel, has comparable performances with the GBF. The improvement is significant at low noise levels.

We validate our results by testing the proposed filters on real endoscopic images. Lesions with irregular margins are indicative of colorectal cancer \cite{kudo} (Fig.~\ref{fig:endo}, Fig.~\ref{fig:endo2}). In the early stages, these lesions are not very significant and might be mistakenly ignored in the presence of noise. For a noisy realization of the image with input PSNR of 18 dB, we observe an improvement of 14 dB using the directional bilateral filter. The improvement with the GBF is 1 dB lesser.

\section{Conclusions} \label{sec:conc}
\vspace{-0.1in}
We have proposed a modified bilateral filter that combines two edge-preservation techniques. The domain kernel incorporates orientation and anisotropy of image structures by means of a structure tensor and smooths perpendicular to dominant orientations. By doing so, the influence of outliers is suppressed while smoothing. When combined with the range kernel, the two kernels assist each other in edge preservation. We chose the optimal parameters of the directional bilateral filter by minimizing the SURE cost. The parameters that minimize SURE have been found to be nearly optimal in the MSE sense. We show that the proposed directional bilateral filter has better denoising performance than the Gaussian bilateral filter. We attribute this to its improved edge-preserving capability. Finding a computationally less expensive version of the algorithm, and subsequently evaluating its performance are potential research problems.

\bibliography{bib_dbf}
\bibliographystyle{IEEEtran}

\end{document}